\documentclass[10pt,twocolumn,letterpaper]{article}
\usepackage[pagenumbers]{cvpr}
\usepackage[T1]{fontenc}
\usepackage{cuted}
\usepackage{placeins}
\usepackage{colortbl}
\usepackage{multirow}
\usepackage{tabularx}
\definecolor{cvprblue}{rgb}{0.21,0.49,0.74}
\usepackage[pagebackref,breaklinks,colorlinks,allcolors=cvprblue]{hyperref}

\def\confName{CVPR}
\def\confYear{2027}

\title{Grounding Across Views: Cross-View Referring Reasoning for Visual Grounding}
\newcommand{\arxivauthornames}{%
  \textbf{Hao Gan\textsuperscript{1}, Liuqian Wang\textsuperscript{1,*}, Zhuping Hu\textsuperscript{1}, Yufei Gao\textsuperscript{1},}\\
  \textbf{Wentao Chao\textsuperscript{1}, Chong Fu\textsuperscript{1}, Shuang Li\textsuperscript{1}, Chengming Liu\textsuperscript{1}, Lei Shi\textsuperscript{1}}}
\newcommand{\arxivaffiliation}{\textsuperscript{1}Zhengzhou University, Henan, China}
\newcommand{\arxivcorrespondence}{\textsuperscript{*}Corresponding Author}
\author{%
  \arxivauthornames\\[2pt]
  \arxivaffiliation\\[2pt]
  {\small\arxivcorrespondence}}
\hypersetup{%
  pdftitle={Grounding Across Views: Cross-View Referring Reasoning for Visual Grounding},
  pdfauthor={Hao Gan, Liuqian Wang, Zhuping Hu, Yufei Gao, Wentao Chao, Chong Fu, Shuang Li, Chengming Liu, Lei Shi}}

\makeatletter
\patchcmd{\@maketitle}{\vskip .375in}{\vskip 4pt}{}{\PackageError{GAV}{Cannot set arXiv title spacing}{}}
\patchcmd{\@maketitle}{\vspace*{12pt}}{\vspace*{0pt}}{}{\PackageError{GAV}{Cannot set arXiv title trailing space}{}}
\patchcmd{\@maketitle}{\vspace*{24pt}}{\vspace*{8pt}}{}{\PackageError{GAV}{Cannot set arXiv author spacing}{}}
\makeatother
\begin{document}
\maketitle
\begin{abstract}
Cross-view object geo-localization (CVOGL) aims to identify a query-view target in a satellite image, yet reliable instance grounding remains challenging under substantial viewpoint discrepancies. Although recent methods incorporate multimodal prompts, they insufficiently address the varying transferability of linguistic and visual evidence across views. Viewpoint-dependent descriptions (e.g., building facades visible in oblique UAV views but not in top-down satellite images) may become unreliable for satellite grounding, while visually similar instances further complicate target identification. To address these challenges, we rethink CVOGL from a visual grounding perspective and propose GAV (Grounding Across Views), a view-aware framework for constructing target-centered evidence and establishing discriminative cross-view correspondence. Specifically, view-aware referring reasoning mitigates viewpoint-dependent ambiguity, multi-token exemplar prompting and contrastive alignment strengthen visual-semantic correspondence, and scene-level supervision with alignment-based verification distinguishes competing satellite instances. We further introduce CVOGL-Text, a street- and drone-to-satellite benchmark with query-derived expressions. GAV achieves state-of-the-art performance on MoP-UAV with 62.55\% Acc@0.5 (+18.19 percentage points) and outperforms evaluated click-point baselines on CVOGL-Text.
\end{abstract}

\begin{figure}[t]
\centering
\includegraphics[width=\linewidth]{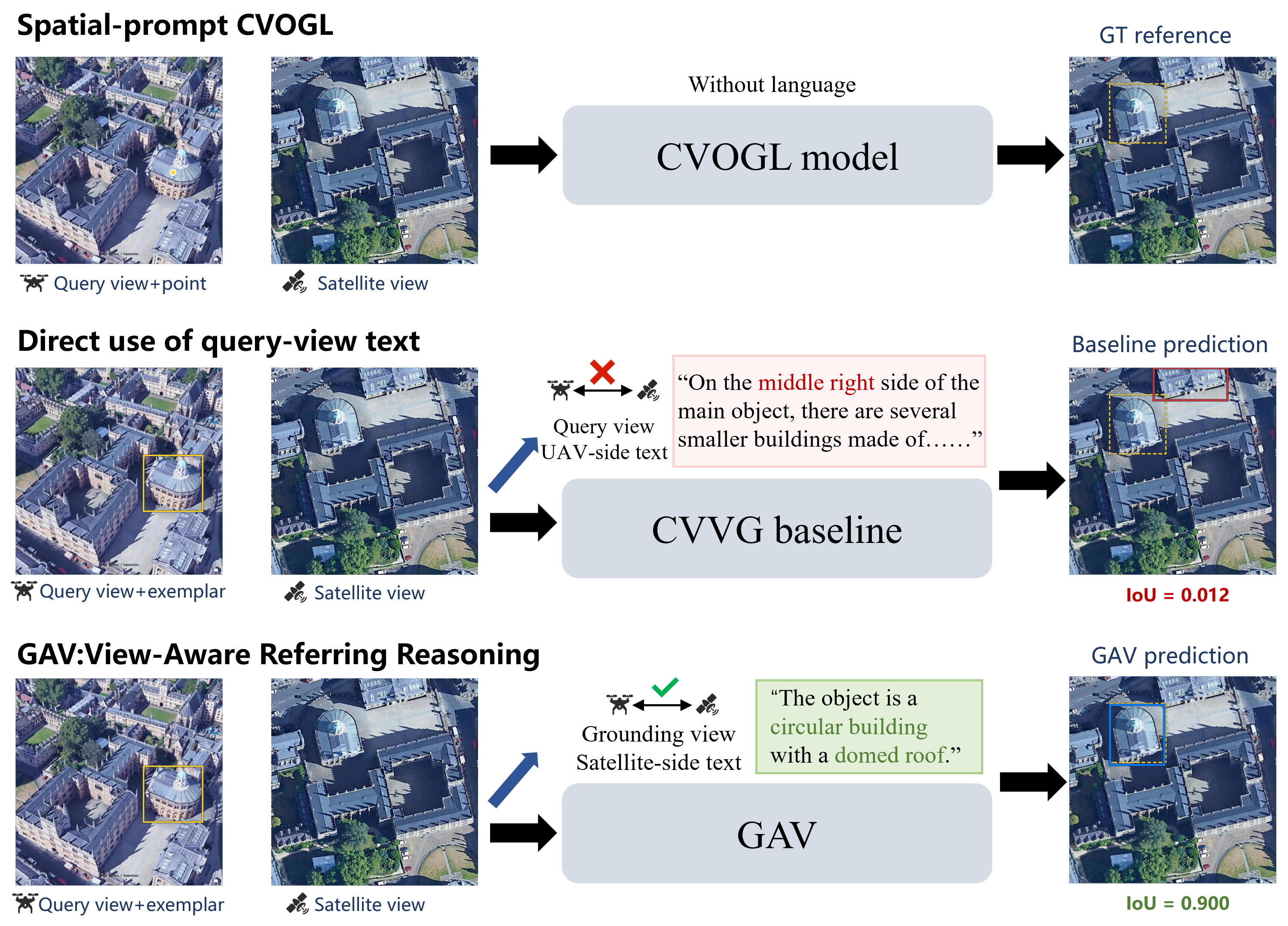}

\caption{Motivation for view-aware cross-view visual grounding. Top: conventional spatial prompting specifies the query target, with satellite ground truth shown as reference. Middle: directly transferring query-view expressions may lead to incorrect satellite-instance grounding. Bottom: GAV constructs target-centered semantic evidence and integrates visual exemplars for reliable cross-view grounding. Predictions are from separately trained configurations on the same sample (IoU 0.012 vs. 0.900), rather than a text-only comparison.}

\label{fig:teaser}
\end{figure}

\section{Introduction}

Cross-view geo-localization establishes spatial correspondence between observations captured from different platforms and viewpoints \citep{liang2026unigeors,song2026geobridge,lv2026mogeo}. It is particularly important when satellite navigation signals are unavailable or unreliable, such as in urban canyons, disaster response, and low-altitude autonomous navigation \citep{zhu2023sues,dai2024denseuav}. While conventional methods typically retrieve a corresponding satellite image or geographic region \citep{zhu2022transgeo,deuser2023sample4geo}, many practical applications require locating a specific object instance within that region. Cross-view object geo-localization (CVOGL) addresses this need by anchoring an object observed from a drone or street-level platform to its precise location in a satellite image \citep{sun2023detgeo}.

Existing CVOGL methods typically designate query targets using a click point or bounding box \citep{sun2023detgeo,huang2025ocgnet}, while recent multimodal approaches additionally exploit natural language \citep{zhang2026mop}. However, language describing an object in the query view is not necessarily reliable for grounding in a satellite image. Although target attributes and contextual cues may remain informative, query-relative directions and positional references cannot be directly transferred across viewpoints. Moreover, satellite scenes often contain multiple visually or semantically similar instances, making it difficult to associate the available evidence with the intended target. The central challenge is therefore not simply incorporating language, but constructing reliable target-specific evidence and associating it with the correct satellite instance, as illustrated in Fig.~\ref{fig:teaser}.

Our key insight is that semantic and visual cues differ in their transferability across viewpoints and should be selectively constructed and integrated rather than uniformly transferred. This motivates three key requirements.
\textbf{(i) View-aware semantic construction.} Query-view descriptions should preserve informative target attributes and contextual relations while avoiding unreliable cross-view interpretations of viewpoint-dependent cues.
\textbf{(ii) Exemplar-guided cross-view association.} Substantial appearance discrepancies across viewpoints require complementary integration of linguistic descriptions and instance-specific visual evidence to establish reliable target correspondence.
\textbf{(iii) Scene-level instance discrimination.} Visually or semantically similar objects within a satellite scene require explicit discrimination between the referred target and competing instances.

To address these requirements, we investigate cross-view visual grounding (CVVG) and propose \textbf{GAV} (\textbf{G}rounding \textbf{A}cross \textbf{V}iews), a unified framework for view-aware semantic construction and discriminative cross-view instance grounding. Specifically, view-aware referring reasoning derives target-centered descriptions from the marked query image, verifies their category consistency against the original expression, and incorporates a coarse direction-free positional role without assuming direct geometric correspondence across views. Multi-token exemplar prompting integrates instance-specific visual evidence into language representations, while cross-view contrastive alignment strengthens correspondence with satellite targets. Finally, scene-level discriminative supervision and alignment-based verification improve target selection among competing instances. To facilitate evaluation across different viewpoints, we further introduce CVOGL-Text, extending street- and drone-view CVOGL data with query-derived referring expressions.

Our main contributions are summarized as follows:
\begin{itemize}
    \item We rethink cross-view object geo-localization from a visual grounding perspective, shifting the focus from spatial target specification to cross-view instance grounding guided by referring expressions and visual exemplars. We further introduce CVOGL-Text, a benchmark for street- and drone-to-satellite grounding with query-derived expressions.

    \item We propose GAV, a view-aware grounding framework that constructs target-centered semantic evidence to mitigate viewpoint-dependent ambiguity. Multi-token exemplar prompting and cross-view contrastive alignment further establish complementary visual-semantic correspondence across viewpoints.

    \item We introduce scene-level discriminative grounding with alignment-based verification to distinguish competing satellite instances. Extensive experiments achieve 62.55\% Acc@0.5 on MoP-UAV (+18.19 percentage points over the previous best) and demonstrate improvements on CVOGL-Text.
\end{itemize}

\section{Related Work}

\textbf{Cross-view object geo-localization.}
DetGeo \citep{sun2023detgeo} introduced clicked-target CVOGL, and OCGNet \citep{huang2025ocgnet} strengthened the click prompt; later work considers multi-object and language-guided settings \citep{lv2026mogeo,wu2026geomatch}. Cross-view retrieval methods match whole images \citep{zheng2020university,zhu2023sues,dai2024denseuav,ji2025game4loc,deuser2023sample4geo,mi2024congeo} but do not localize instances. MoP-UAV \citep{zhang2026mop}, the closest setting, pairs UAV--satellite imagery with expressions. Unlike task-specific multimodal prompting, CVVG studies which linguistic and exemplar evidence remains transferable after viewpoint change and adapts a grounding foundation model accordingly.

\textbf{Remote-sensing object recognition.}
Position-guided receptive fields improve small-object detection~\citep{Wang2025PGDRFNet}, while rotation-equivariant features support oriented detection~\citep{Ming2026HERO}. Fine-grained recognition uses knowledge distillation and filter grafting~\citep{Wang2023FilterGrafting} or VLM-guided regional restoration~\citep{Wang2026WeakSupervision}. These methods address within-image recognition; CVVG additionally requires associating a specified instance across viewpoints.

\textbf{Open-vocabulary detection and visual grounding.}
Vision-language grounding builds on contrastive pretraining and DETR-style detection \citep{radford2021clip,carion2020detr,zhang2023dino,li2022glip,kamath2021mdetr,liu2024gdino} to align region and text features, while multimodal LLMs extend grounding to open-ended instructions \citep{bai2025qwen3vl}. Token-aggregated region scores link expression quality to candidate ranking. However, these models assume that the expression and target image share a viewpoint, which CVVG violates.

\textbf{Referring expression comprehension.}
RefCOCO and referring models study within-image localization \citep{yu2016refcoco,mao2016generation,yu2018mattnet,deng2021transvg}. GeoText-1652 brings expressions to aerial imagery \citep{chu2024geotext}, language-guided street localization remains image-level retrieval \citep{ye2025cvgtext}, and remote-sensing grounding stays within a single nadir view \citep{zhan2023rsvg}. CVVG instead grounds an expression authored from another viewpoint, making cue transferability central.

\section{Methodology}

\setcounter{dbltopnumber}{1}

\begin{figure*}[t]
\centering
\includegraphics[width=\textwidth]{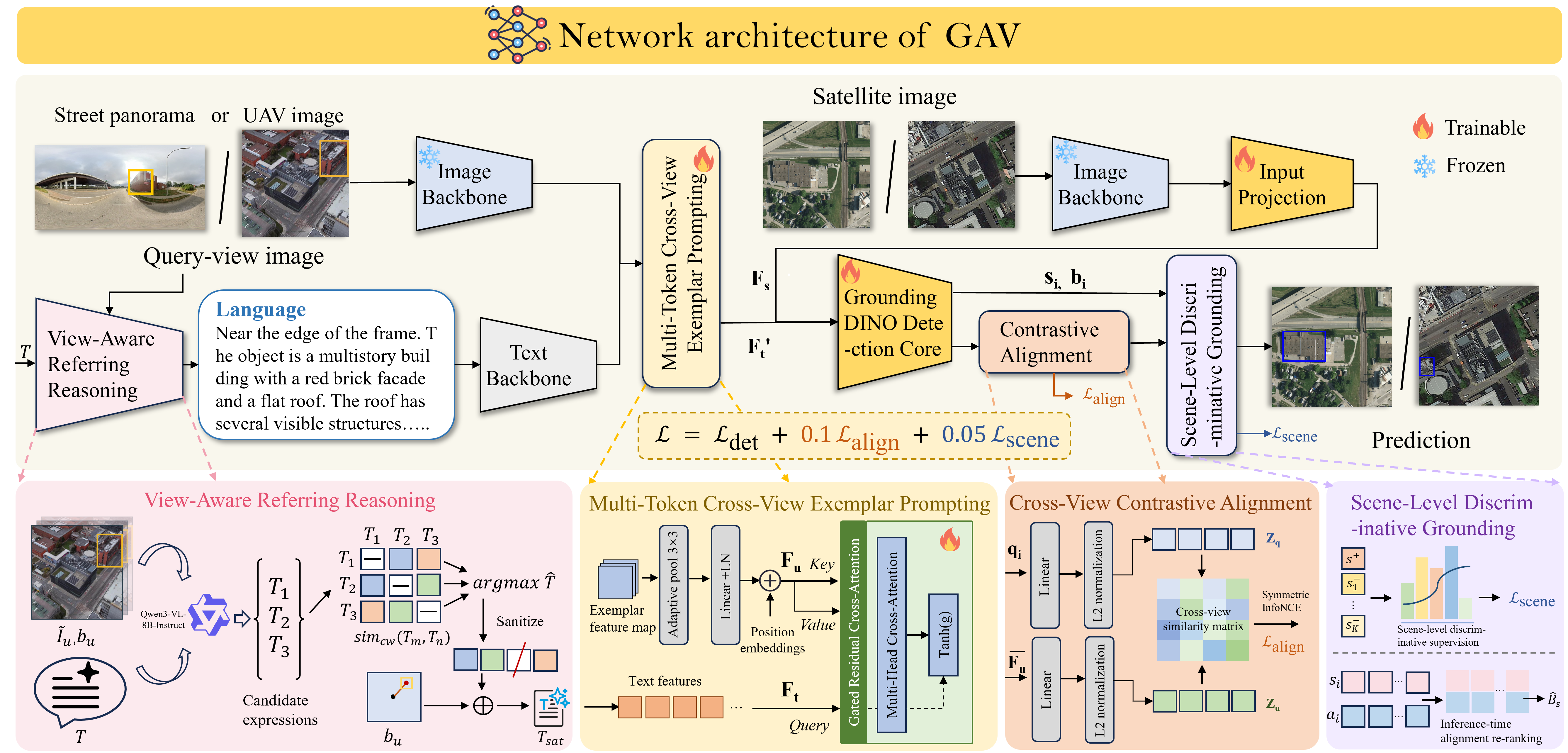}
\caption{Overview of the proposed GAV framework. View-aware referring reasoning constructs target-centered descriptions from the marked query image, using the original expression only for category checking and fallback. Multi-token exemplar prompting incorporates instance-specific visual evidence, while cross-view contrastive alignment strengthens exemplar-to-satellite correspondence. Scene-level discriminative supervision distinguishes competing instances, followed by alignment-based re-ranking for final target selection. Satellite ground-truth boxes are used only during training.}
\label{fig:arch}
\end{figure*}

\subsection{Task Definition and CVOGL-Text}
\label{sec:task}

\textbf{Task definition.}
Cross-view visual grounding localizes a query-view target in a satellite image. Given the satellite image $I_s$, query image $I_u$, exemplar box $b_u$, and referring expression $T$, the complete pipeline predicts
\begin{equation}
    \hat{B}_s=f_{\theta}(I_s,I_u,b_u,T).
\label{eq:task}
\end{equation}
The exemplar identifies the query instance, while language describes its attributes and context. GAV derives a satellite-groundable expression $T_{\mathrm{sat}}$ offline from the marked query image; source $T$ supports category checking and conditional fallback. The online localizer receives $I_s$, the exemplar, and $T_{\mathrm{sat}}$. Neither expression generation nor inference accesses the satellite ground-truth box.

\textbf{CVOGL-Text.}
We extend CVOGL~\citep{sun2023detgeo} with referring expressions for text-guided street/drone-to-satellite grounding (Table~\ref{tab:benchcmp}). CVOGL-Text retains the 19 categories and official 4{,}343/923/973 training/validation/test samples per query modality, sharing the original satellite pool. Each click defines a square exemplar of side $0.3\min(H,W)$. A frozen Qwen3-VL-8B-Instruct~\citep{bai2025qwen3vl} generates descriptions from the marked query image and a fixed ground-truth category constraint, without the satellite image, satellite box, or category-only seed text. Three candidates undergo consensus selection, sanitization, category checking, and role-cue concatenation; the seed supplies only the category check and fallback. Five annotators screen sampled test expressions for category correctness, query-visible fidelity, and groundability, and flagged cases are regenerated. Expressions are cached independently of grounding-model training. The same query-only procedure, without a ground-truth category constraint, also derives expressions for MoP-UAV~\citep{zhang2026mop}. The Appendix gives construction details.

\begin{table}[t]
\caption{Comparison with cross-view geo-localization benchmarks. IGL and OGL denote image- and object-level geo-localization, respectively. Columns marked $\dagger$ are contributed by this work, and GeoText-1652 \citep{chu2024geotext} counts follow \citet{zhang2026mop}.}
\label{tab:benchcmp}
\centering
\footnotesize
\setlength{\tabcolsep}{2.5pt}
\renewcommand{\arraystretch}{1.02}
\begin{tabular}{l ccc >{\columncolor{gray!12}}c >{\columncolor{gray!12}}c}
\toprule
 & \shortstack{GeoText\\-1652} & CVOGL & \shortstack{MoP\\-UAV} & \shortstack{MoP\\-UAV$^{\dagger}$} & \shortstack{CVOGL\\-Text$^{\dagger}$} \\
\midrule
Task & IGL & OGL & OGL & OGL & OGL \\
Query view & UAV & \shortstack{drone,\\street} & UAV & UAV & \shortstack{drone,\\street} \\
Point & -- & \checkmark & \checkmark & \checkmark & \checkmark \\
Box & \checkmark & -- & \checkmark & \checkmark & -- \\
Language & template & -- & template & \textbf{derived} & \textbf{derived} \\
Categories & -- & 19 & -- & -- & \textbf{19} \\
Expressions & 113.3k & -- & 102.9k & \textbf{102.9k} & \textbf{12.5k} \\
\bottomrule
\end{tabular}

\end{table}

\subsection{Overview}
Figure~\ref{fig:arch} summarizes GAV. Referring reasoning derives $T_{\mathrm{sat}}$; exemplar prompting conditions its features on the query target. Alignment and scene supervision distinguish satellite candidates, followed by alignment re-ranking.

Following Grounding DINO~\citep{liu2024gdino}, text tokens $F_t=\{t_j\}_{j=1}^{L}$ interact with $N$ satellite decoder queries $\{q_i\}_{i=1}^{N}$, each predicting a box $\hat b_i$. For content tokens $\mathcal P$, the standard score and prediction are
\begin{equation}
 s_i=\frac{1}{|\mathcal P|}\sum_{j\in\mathcal P}\sigma(q_i^\top t_j),
 \qquad \hat B_s=\hat b_{\arg\max_i s_i}.
\label{eq:aggscore}
\end{equation}

\begin{figure*}[t]
\centering
\includegraphics[width=\textwidth]{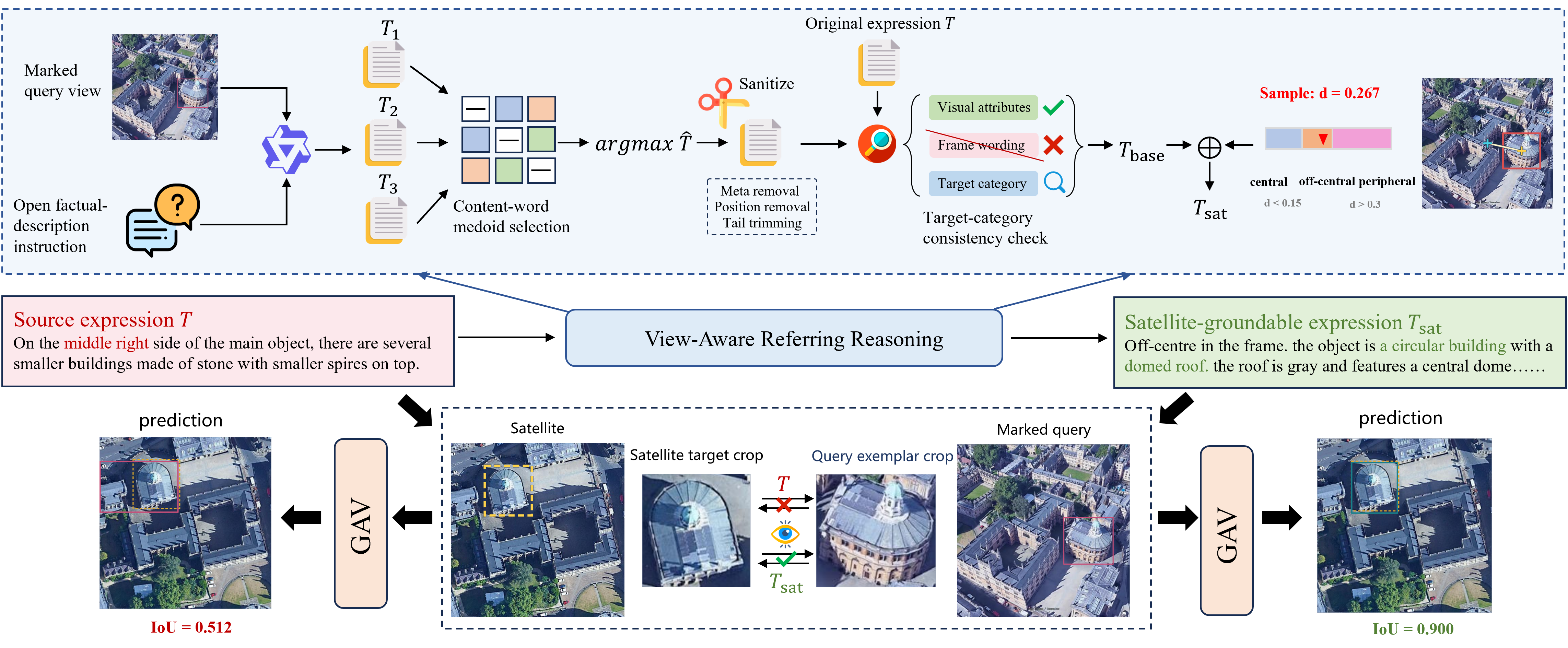}
\caption{View-aware referring reasoning. Top: query-only VLM generation, content-word medoid selection, sanitization, and a category check against source $T$ produce $T_{\mathrm{base}}$; a direction-free query-frame role yields $T_{\mathrm{sat}}$. The illustrated centroid distance $d=0.267$ gives an off-center role. Source $T$ supplies the category check and conditional fallback, not the generative prompt. Bottom: with GAV weights, images, and exemplar fixed, $T$ and $T_{\mathrm{sat}}$ yield IoU 0.512 and 0.900 ($\lambda=0$). Target crops and satellite annotations illustrate correspondence only. This is a paired example, not an aggregate consistency score.}
\label{fig:varr}
\end{figure*}

\subsection{View-Aware Referring Reasoning}

We distinguish intrinsic attributes, query-frame directions, and spatial-layout cues by their expected cross-view transferability. GAV generates factual descriptions from the marked query image, removes image-position wording, and adds a coarse direction-free role (Fig.~\ref{fig:varr}). It operates on generated text rather than editing $T$.

\textbf{Expression derivation.}
Qwen3-VL-8B-Instruct \citep{bai2025qwen3vl} receives the marked query view $\widetilde I_u$ and an instruction requesting visible appearance, shape, and material. The prompt excludes $T$ and the satellite image; CVOGL-Text additionally supplies a fixed category constraint. From three candidates $\mathcal{T}=\{T_m\}_{m=1}^{3}$, we select the content-word medoid and sanitize it:
\begin{equation}
\begin{aligned}
\widehat{T}
&=\arg\max_{T_m\in\mathcal{T}}
\sum_{\substack{n=1\\n\ne m}}^{3}
\operatorname{sim}_{\mathrm{cw}}(T_m,T_n),\\
T_{\mathrm{base}}&=\mathcal{S}(\widehat{T};T).
\end{aligned}
\end{equation}

Here, $\operatorname{sim}_{\mathrm{cw}}$ is the Jaccard overlap of content-word sets; medoid selection retains one complete candidate rather than merging fragments. The sanitizer removes annotation and image-position meta-phrases and trims incomplete outputs. It also compares automatically identified target categories in $\widehat{T}$ and $T$, falling back to $T$ only when both categories are identifiable and disagree.

\textbf{Positional cue.}
Let $r$ be the normalized distance between the exemplar centroid $p(b_u)$ and the image center $c(I_u)$. We prepend a direction-free role phrase:
\begin{equation}
\begin{aligned}
r &= d_{\mathrm{norm}}\!\left(p(b_u),c(I_u)\right),\\
\pi(r)&=
\begin{cases}
\text{central}, & r<0.15,\\
\text{off-center}, & 0.15\le r<0.30,\\
\text{peripheral}, & r\ge 0.30,
\end{cases}\\
T_{\mathrm{sat}}&=\pi(r)\oplus T_{\mathrm{base}},
\end{aligned}
\end{equation}
where $\oplus$ denotes phrase concatenation. This phrase encodes a coarse role in the query frame rather than an assumed geometric correspondence between query and satellite coordinates. Figure~\ref{fig:varr} denotes this distance by $d$; its example has $d=0.267$ and receives the off-center role.

The cached $T_{\mathrm{sat}}$ conditions online grounding. In Fig.~\ref{fig:varr}, replacing $T$ with $T_{\mathrm{sat}}$ raises IoU from 0.512 to 0.900 with weights and visual inputs fixed and no re-ranking. The Appendix provides attention and response maps; satellite target crops are visualization references only.

\subsection{Multi-Token Exemplar Prompting}

To supply visual instance evidence, we encode the query-view exemplar as multiple spatial tokens. The region $b_u$ is cropped from $I_u$, resized to $224\times224$, and encoded by the shared frozen image backbone. Adaptive average pooling produces a $3\times3$ grid, and a linear projection, layer normalization, and learned position embeddings yield $F_u\in\mathbb{R}^{9\times d}$, with $d{=}256$.

Text tokens $F_t\in\mathbb{R}^{L\times d}$ attend to $F_u$ through one eight-head gated residual cross-attention layer,
\begin{equation}
F_t' \;=\; F_t + \tanh(g)\cdot \mathrm{LN}\!\left(\mathrm{MHA}\left(F_t,\, F_u,\, F_u\right)\right),
\label{eq:gate}
\end{equation}
The attention queries and key/value inputs are layer-normalized; $F_t'$ retains shape $L\times d$ and is passed to the grounding core. The scalar gate $g$ is initialized to zero, so Equation~\ref{eq:gate} initially preserves the pretrained representation. Exemplar features are injected only into the language branch; early fusion with satellite queries underperforms the text-only base in the ablation.

\subsection{Cross-View Contrastive Alignment}

We align the query-view exemplar with the target-matched satellite query during training.

Let $i^{+}$ denote the decoder query matched to the ground-truth box. Let $\bar F_u=\frac{1}{9}\sum_{j=1}^{9}F_{u,j}$. Each projection $\phi_q,\phi_u$ applies a learned $256\!\to\!256$ linear map followed by $\ell_2$ normalization, yielding $z_q=\phi_q(q_{i^{+}})$ and $z_u=\phi_u(\bar{F}_u)$. Stacking a batch gives the cross-view similarity matrix $S=Z_q Z_u^{\top}/\tau_a$, whose diagonal contains matched pairs. We optimize a symmetric InfoNCE loss \citep{oord2018infonce} with in-batch negatives $\mathcal{B}$ and temperature $\tau_a{=}0.07$,
\begin{equation}
\begin{aligned}
\mathcal{L}_{\mathrm{align}} = -\tfrac{1}{2}\Big[
&\log \tfrac{\exp(z_q^{\top} z_u/\tau_a)}{\sum_{u' \in \mathcal{B}} \exp(z_q^{\top} z_{u'}/\tau_a)} \\[1pt]
{}+{}&\log \tfrac{\exp(z_u^{\top} z_q/\tau_a)}{\sum_{q' \in \mathcal{B}} \exp(z_u^{\top} z_{q'}/\tau_a)}
\Big].
\end{aligned}
\end{equation}

With weight 0.1, $\mathcal{L}_{\mathrm{align}}$ pulls matched query--exemplar pairs together and separates in-batch instances, discouraging same-category mismatches. Projection heads are bypassed in the standard decision and reused for re-ranking.

\subsection{Scene-Level Supervision}

In-batch negatives may omit competing satellite objects. Scene supervision directly compares grounding scores of the referred instance and annotated in-scene alternatives under the same expression, complementing cross-view exemplar alignment.

For each training sample, $s^{+}$ is the score in Equation~\ref{eq:aggscore} for the query matched to the referred target, and $\{s^{-}_k\}_{k=1}^{K}$ are scores of queries assigned to other annotated targets in the same scene. Competitor queries must overlap their targets by at least 0.25 IoU, while boxes overlapping the referred target above 0.5 IoU are excluded. Competitors are drawn from all other annotated targets satisfying these constraints and are not restricted by category. We optimize
\begin{equation}
\mathcal{L}_{\mathrm{scene}} = -\log
\frac{\exp(s^{+}/\tau_s)}{\exp(s^{+}/\tau_s) + \sum_{k=1}^{K} \exp(s^{-}_k/\tau_s)},
\end{equation}
with $\tau_s{=}0.05$ and at most $K{=}8$ competitors. This loss adds no learnable parameters and follows soft set-level ranking objectives \citep{wang2019multisimilarity}.

\subsection{Alignment Re-ranking}
At inference, we reuse the alignment head to verify the top $K_r{=}5$ candidates selected by text score. Let $\tilde{s}_i$ and $\tilde{a}_i$ denote text and alignment scores min-max normalized over all decoder queries, with $a_i$ given by the cosine similarity between $\phi_q(q_i)$ and $\phi_u(\bar{F}_u)$. The final prediction is
\begin{equation}
i^{*} = \arg\max_{i \in \mathcal{C}} \;(1-\lambda)\, \tilde{s}_i + \lambda\, \tilde{a}_i ,
\end{equation}
where $\mathcal{C}$ is the candidate set and $\lambda$ is selected on held-out training data for MoP-UAV and on the validation split for CVOGL-Text. Setting $\lambda{=}0$ recovers the standard decision.

\textbf{Training objective.}
The optimization objective is
$\mathcal{L} = \mathcal{L}_{\mathrm{det}} + w_a \mathcal{L}_{\mathrm{align}} + w_s \mathcal{L}_{\mathrm{scene}}$,
where $\mathcal{L}_{\mathrm{det}}$ combines focal classification \citep{lin2017focal}, $\ell_1$, and GIoU \citep{rezatofighi2019giou} terms, $w_a{=}0.1$, and $w_s{=}0.05$.

\section{Experiments}

\subsection{Experimental Setup}
We evaluate on MoP-UAV~\citep{zhang2026mop}, with 85{,}469/17{,}447 training/test triplets, and CVOGL-Text (Sec.~\ref{sec:task}). We report Acc@0.25, Acc@0.5, and mean IoU; Acc@$\tau$ is the fraction of predictions with IoU above $\tau$. Swin-T~\citep{liu2021swin} and BERT~\citep{devlin2019bert} remain frozen. Training lasts 12 epochs on MoP-UAV and 24 on CVOGL-Text, and evaluation uses the final checkpoint. Expressions are precomputed, so no VLM runs during grounding inference. The Appendix gives the training recipe, hardware, and model-selection protocol.

\begin{figure*}[t]
\centering
\input{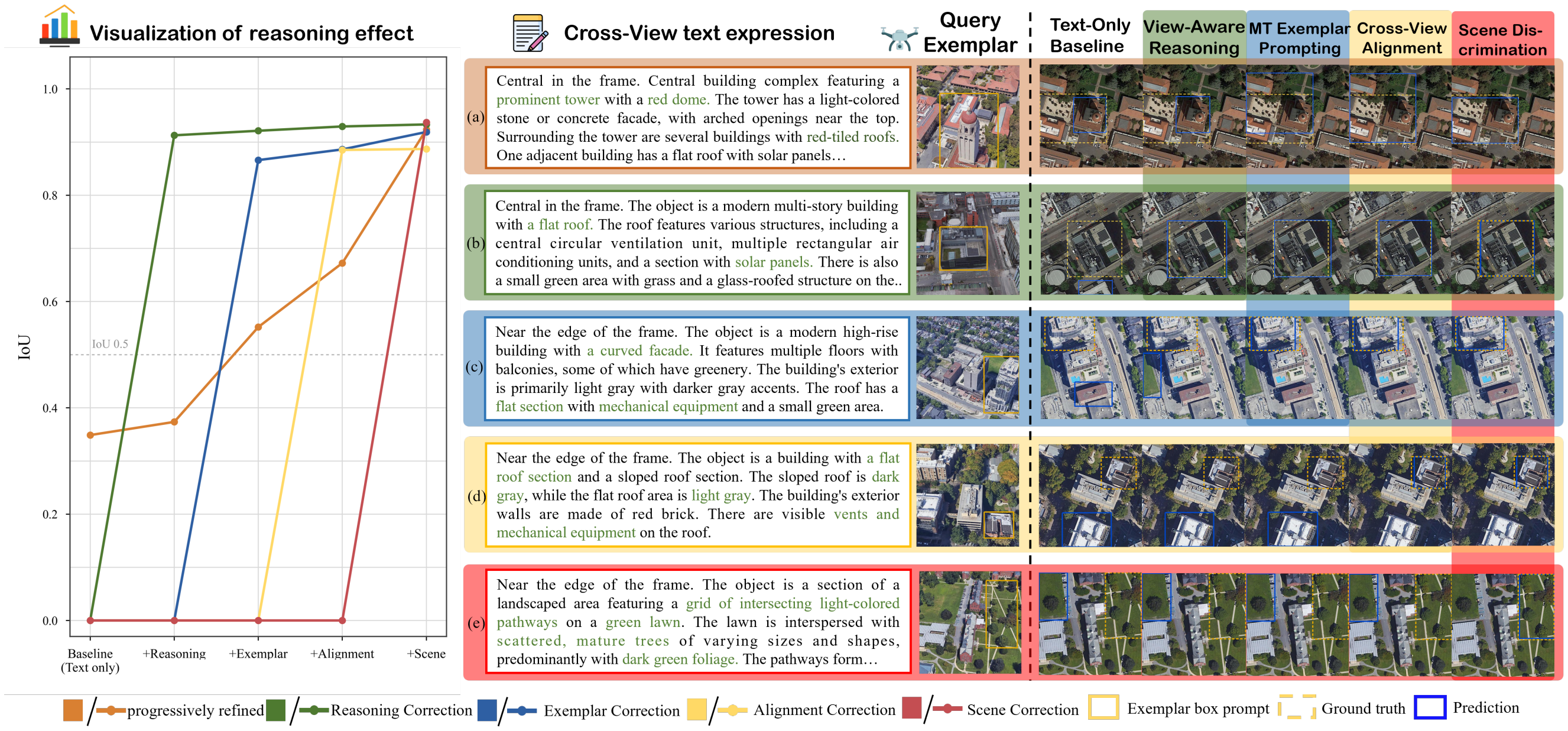}
\caption{Cumulative predictions on MoP-UAV: (a, orange) gradual refinement; (b, green) correction after referring reasoning; (c, blue) after exemplar prompting; (d, yellow) after alignment; (e, red) after scene supervision and re-ranking. Columns add each method to the preceding configuration; curves show IoU. Gold solid/dashed boxes mark query exemplars/satellite ground truth, and blue boxes mark predictions. Text cards show $T_{\mathrm{sat}}$ excerpts with selected visual cues in green; the baseline uses $T$. Each configuration is trained separately. The final stage includes top-5 re-ranking at $\lambda=0.5$.}
\label{fig:qual}
\end{figure*}

\begin{figure*}[t]
\centering
\begin{minipage}[b]{0.51\textwidth}
\centering
\includegraphics[width=\linewidth]{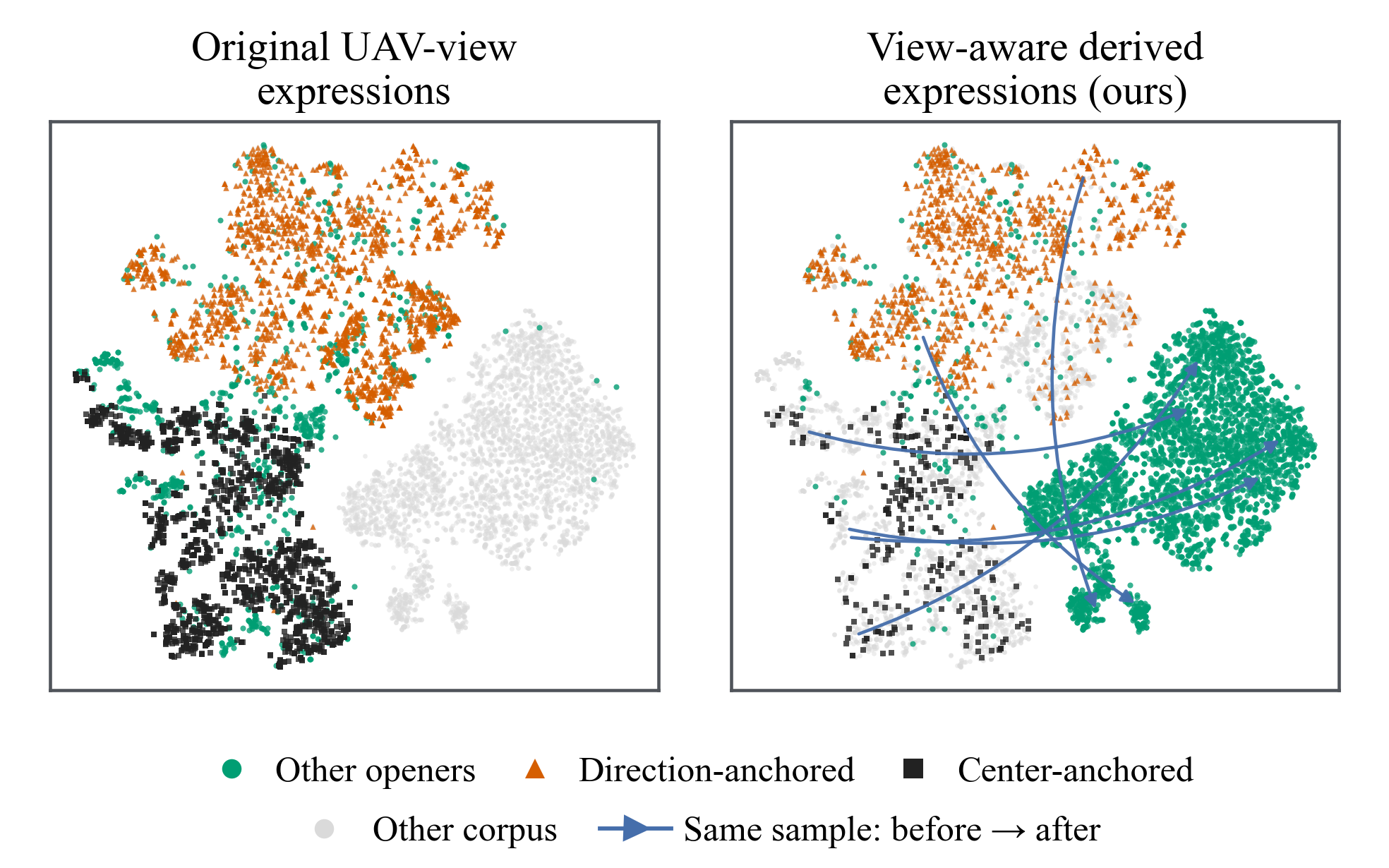}\\[-1pt]
{\small (a) Original and derived expression embeddings}
\end{minipage}\hfill
\begin{minipage}[b]{0.47\textwidth}
\centering
\includegraphics[width=\linewidth]{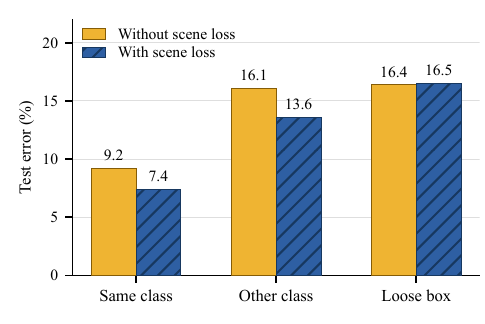}\\[-1pt]
{\small (b) Effect of scene supervision}
\end{minipage}
\caption{Text and instance analysis on MoP-UAV. (a) Joint t-SNE of 5,000 paired frozen-BERT embeddings before role-cue concatenation. Panels share one space; gray points show the other corpus. Colors group expressions by their opening phrase, and arrows link six selected pairs. (b) Test error rates without/with scene supervision, before re-ranking. Same-/other-class selection errors decrease, while loose boxes ($0.25\leq\mathrm{IoU}<0.5$ on the correct instance) remain nearly unchanged.}
\label{fig:analysis}
\end{figure*}

\begin{table}[!tb]
\caption{Comparison with state-of-the-art methods on the MoP-UAV test set. Baselines are reported by \citet{zhang2026mop} on the same split. GAV uses text and a query-view exemplar without a point prompt; its re-ranking weight is selected on held-out training data.}
\label{tab:main}
\renewcommand{\arraystretch}{1.08}
\centering\small
\setlength{\tabcolsep}{2.5pt}
\begin{tabularx}{\linewidth}{@{}>{\raggedright\arraybackslash}Xrrr@{}}
\toprule
Method / prompt & Acc@0.25 & Acc@0.5 & mIoU \\
\midrule
DetGeo \citep{sun2023detgeo}, point prompt & 0.4830 & 0.3208 & 0.3273 \\
OCGNet \citep{huang2025ocgnet}, point prompt & 0.4970 & 0.3349 & 0.3373 \\
MoPT \citep{zhang2026mop}, point prompt & 0.5983 & 0.4051 & 0.3685 \\
MoPT \citep{zhang2026mop}, text + box prompts & 0.6277 & 0.4387 & 0.3846 \\
MoPT \citep{zhang2026mop}, text + box + point & 0.6366 & 0.4436 & 0.3917 \\
\midrule
GAV (no re-ranking) & 0.7863 & 0.6209 & 0.5597 \\
GAV & \textbf{0.7901} & \textbf{0.6255} & \textbf{0.5625} \\
\bottomrule
\end{tabularx}
\end{table}

\begin{table}[!t]
\caption{Results on CVOGL-Text, with 973 test samples per query view. GAV uses $\lambda{=}0.7$, selected on validation data. Dashes indicate unreported mIoU.}
\label{tab:cvogl}
\renewcommand{\arraystretch}{1.0}
\centering
\small
\setlength{\tabcolsep}{2pt}
\begin{tabularx}{\linewidth}{@{}>{\raggedright\arraybackslash}Xrrr@{}}
\toprule
Method & Acc@0.25 & Acc@0.5 & mIoU \\
\midrule
\multicolumn{4}{@{}l}{\emph{Street-view queries}} \\
CVM-Net \citep{hu2018cvmnet} & 0.0473 & 0.0051 & -- \\
TransGeo \citep{zhu2022transgeo} & 0.2117 & 0.0288 & -- \\
GeoDTR \citep{zhang2023geodtr} & 0.3237 & 0.0606 & -- \\
Sample4Geo \citep{deuser2023sample4geo} & 0.0884 & 0.0123 & -- \\
DetGeo \citep{sun2023detgeo} & 0.4543 & 0.4224 & -- \\
VAGeo \citep{li2025vageo} & 0.4821 & 0.4522 & -- \\
MOGeo \citep{lv2026mogeo} & 0.5098 & 0.4770 & -- \\
Category text only & 0.3751 & 0.3412 & 0.2969 \\
\quad + exemplar, alignment & 0.3299 & 0.2991 & 0.2610 \\
Derived text only & 0.5118 & 0.4769 & 0.4039 \\
GAV (no re-ranking) & 0.5087 & 0.4635 & 0.3980 \\
GAV & \textbf{0.5252} & \textbf{0.4800} & \textbf{0.4099} \\
\midrule
\multicolumn{4}{@{}l}{\emph{Drone-view queries}} \\
CVM-Net \citep{hu2018cvmnet} & 0.2014 & 0.0329 & -- \\
TransGeo \citep{zhu2022transgeo} & 0.3505 & 0.0637 & -- \\
GeoDTR \citep{zhang2023geodtr} & 0.3248 & 0.0565 & -- \\
Sample4Geo \citep{deuser2023sample4geo} & 0.0771 & 0.0092 & -- \\
DetGeo \citep{sun2023detgeo} & 0.6197 & 0.5766 & -- \\
VAGeo \citep{li2025vageo} & 0.6619 & 0.6187 & -- \\
MOGeo \citep{lv2026mogeo} & 0.6639 & 0.5971 & -- \\
Derived text only & 0.5797 & 0.5231 & 0.4504 \\
GAV (no re-ranking) & 0.6166 & 0.5642 & 0.4835 \\
GAV & \textbf{0.6876} & \textbf{0.6362} & \textbf{0.5380} \\
\bottomrule
\end{tabularx}

\end{table}

\subsection{Comparison with Prior Methods}
\textbf{MoP-UAV.}
GAV reaches 0.6255 Acc@0.5 (Table~\ref{tab:main}), exceeding the previous best by 18.2 points and modality-matched MoPT (text + box) by 18.7 points, without a point prompt. The fine-tuned text-only base reaches 0.4104 (Table~\ref{tab:chain}), below the prior best of 0.4436; the gain cannot be attributed to the grounding backbone alone.

\textbf{CVOGL-Text.}
GAV reaches 0.4800/0.6362 Acc@0.5 for street/drone queries (Table~\ref{tab:cvogl}) and exceeds the strongest listed click-point method by 1.5/2.4 Acc@0.25 points. With derived text, the visual configuration without re-ranking changes Acc@0.25 from 0.5118 to 0.5087 for street views, but from 0.5797 to 0.6166 for drone views. Category text with exemplar and alignment instead loses 4.5 points on street views. Visual evidence therefore depends on both the viewpoint gap and accompanying language. Re-ranking adds 1.65/7.1 Acc@0.25 points for street/drone views.

\subsection{Qualitative Results}
Figure~\ref{fig:qual} follows the method order. The tower complex in (a) improves from IoU 0.349 to 0.924. Samples (b)--(e), a flat-roofed building, a curved high-rise, a building with contrasting roof sections, and a lawn with intersecting paths, first exceed IoU 0.5 after the respective additions. Green phrases mark visual correspondences, not attention weights. These selected cases do not imply monotonic gains on every sample; the Appendix reports full-test corrections and regressions.

The final addition corrects 1,733 test samples at IoU 0.5 and regresses on 932. Only 2,502 of 17,447 samples have nondecreasing IoU across all five configurations, including unchanged failures. Table~\ref{tab:chain} reports the net gains.

\begin{table}[!tb]
\caption{Text perturbations on MoP-UAV with the frozen exemplar-prompted model and original captions. $\Delta$ denotes Acc@0.5 change in percentage points.}
\label{tab:diag}
\renewcommand{\arraystretch}{1.08}
\centering\small
\begin{tabularx}{\linewidth}{@{}Xrr@{}}
\toprule
Text input & Acc@0.5 & $\Delta$ \\
\midrule
Original & 0.4651 & -- \\
Direction swap & 0.4655 & $+0.04$ \\
Layout removal & 0.4480 & $-1.71$ \\
\bottomrule
\end{tabularx}
\end{table}

\begin{table}[!tb]
\caption{Text controls with fixed GAV weights on all 17{,}447 MoP-UAV test samples (Table~\ref{tab:chain} runtime). Mismatches use other scenes within rule-inferred categories (three-seed mean).}
\label{tab:text_controls}
\renewcommand{\arraystretch}{1.08}
\centering\small
\setlength{\tabcolsep}{2.5pt}
\begin{tabularx}{\linewidth}{@{}>{\raggedright\arraybackslash}Xrrr@{}}
\toprule
Text input & Acc@0.25 & Acc@0.5 & mIoU \\
\midrule
Category & 0.5595 & 0.3422 & 0.3574 \\
\quad + role cue & 0.7582 & 0.5837 & 0.5302 \\
Original $T$ & 0.6975 & 0.4793 & 0.4556 \\
Derived $T_{\mathrm{sat}}$ & \textbf{0.7898} & \textbf{0.6251} & \textbf{0.5626} \\
Mismatched $T$ & 0.6224 & 0.4029 & 0.3994 \\
Mismatched $T_{\mathrm{sat}}$ & 0.7616 & 0.5906 & 0.5362 \\
Neutral: \texttt{object.} & 0.4359 & 0.2265 & 0.2721 \\
\bottomrule
\end{tabularx}
\end{table}

\begin{table}[!tb]
\caption{Effect of scene supervision on MoP-UAV, before re-ranking. Small targets are the lowest area quartile; spatial queries contain at least three spatial phrases.}
\label{tab:scene}
\renewcommand{\arraystretch}{1.08}
\centering\small
\setlength{\tabcolsep}{4pt}
\begin{tabularx}{\linewidth}{@{}Xrr@{}}
\toprule
Measure & Without $\mathcal L_{\mathrm{scene}}$ & With $\mathcal L_{\mathrm{scene}}$ \\
\midrule
\multicolumn{3}{@{}l}{\emph{Test error (\%) $\downarrow$}} \\
Wrong instance & 25.6 & 21.4 \\
\quad Same class & 9.2 & 7.4 \\
\quad Other class & 16.1 & 13.6 \\
Loose box & 16.4 & 16.5 \\
\midrule
\multicolumn{3}{@{}l}{\emph{Subset Acc@0.5 $\uparrow$}} \\
Small targets & 0.372 & 0.423 \\
Spatial queries & 0.497 & 0.551 \\
\bottomrule
\end{tabularx}
\end{table}

\subsection{Text and Instance Analysis}
\textbf{Effect of text cues.}
In Table~\ref{tab:diag}, swapping directions changes Acc@0.5 by 0.04 points, while removing spatial-layout words costs 1.71 points. The frozen exemplar-prompted model is more sensitive to these layout cues, motivating factual descriptions with a coarse direction-free role.

\textbf{Text controls with fixed weights.}
Table~\ref{tab:text_controls} varies only the final checkpoint's text, holding images, exemplars, and re-ranking fixed. Derived text gains 14.58 Acc@0.5 points over $T$. Role cues raise category-only accuracy from 0.3422 to 0.5837; matched descriptions add 4.14 points. Other-scene, same-category derived text loses 3.44 points with the recipient role retained (95\% scene-clustered interval: 2.73--4.19). Neutral text retains the language branch. These controls test grounding inputs; source $T$ only guides category checking and fallback during derivation. The Appendix reports all seeds and both selection rules.

Without ground-truth category labels, we infer eight coarse categories from MoP-UAV text, using \texttt{object.} for 1,573 unrecognized source expressions. Mismatches break exemplar correspondence but preserve category, recipient role, and potentially valid attributes; they do not isolate individual words or establish compositional reasoning.

\textbf{Expression embeddings.}
Joint t-SNE~\citep{maaten2008tsne} of mean-pooled BERT features shows a shift from direction/center-anchored openers toward target descriptions (Fig.~\ref{fig:analysis}a). Colors encode opener rules, not object classes. This corpus pattern complements the paired example in Fig.~\ref{fig:varr}; embedding distances do not measure cross-view correspondence or causal accuracy gains.

\textbf{Instance selection.}
With text and exemplar design fixed, scene supervision reduces wrong-instance errors from 25.6\% to 21.4\% before re-ranking, with little change in loose boxes (Table~\ref{tab:scene}, Fig.~\ref{fig:analysis}b). Same-/other-class confusions fall by 1.8/2.5 points; totals include residual low-overlap or background cases. Acc@0.5 rises from 0.372 to 0.423 for small targets and from 0.497 to 0.551 for spatial queries, supporting improved candidate selection rather than box refinement.

\begin{table}[!t]
\caption{Cumulative ablation in method order on all 17{,}447 MoP-UAV test samples, using one fixed-seed run per configuration and a shared evaluation runtime. Original benchmark and design-ablation records are retained separately (Appendix).}
\label{tab:chain}
\renewcommand{\arraystretch}{1.08}
\centering\small
\setlength{\tabcolsep}{2.5pt}
\begin{tabularx}{\linewidth}{@{}>{\raggedright\arraybackslash}Xrrr@{}}
\toprule
Configuration & Acc@0.25 & Acc@0.5 & mIoU \\
\midrule
Text-only baseline & 0.5814 & 0.4104 & 0.4020 \\
+ referring reasoning & 0.6789 & 0.5105 & 0.4774 \\
+ multi-token exemplar & 0.7033 & 0.5387 & 0.4954 \\
+ cross-view alignment & 0.7443 & 0.5792 & 0.5281 \\
+ scene loss + re-ranking & \textbf{0.7898} & \textbf{0.6251} & \textbf{0.5626} \\
\bottomrule
\end{tabularx}
\end{table}

\begin{table}[!t]
\caption{Exemplar-prompt design ablations on the MoP-UAV test set. Fusion variants use the text-only base; token-count variants keep cross-view alignment fixed.}
\label{tab:abl}
\renewcommand{\arraystretch}{1.08}
\centering
\small
\setlength{\tabcolsep}{3.5pt}
\begin{tabular}{@{}lccc@{}}
\toprule
Variant & Acc@0.25 & Acc@0.5 & mIoU \\
\midrule
\multicolumn{4}{@{}l}{\emph{Fusion on the text-only base}} \\
Text-only base & 0.5807 & 0.4100 & 0.4015 \\
Early satellite fusion & 0.5715 & 0.4044 & 0.3986 \\
Additive fusion & 0.6232 & 0.4580 & 0.4362 \\
Gated cross-attention & \textbf{0.6329} & \textbf{0.4651} & \textbf{0.4407} \\
\midrule
\multicolumn{4}{@{}l}{\emph{Token count with alignment}} \\
1 token (global vector) & 0.6656 & 0.5091 & 0.4720 \\
9 tokens ($3\times3$, ours) & \textbf{0.6809} & \textbf{0.5159} & \textbf{0.4794} \\
49 tokens ($7\times7$) & 0.6637 & 0.4907 & 0.4632 \\
\bottomrule
\end{tabular}

\end{table}

\begin{table}[!t]
\caption{Alignment re-ranking ablations on the MoP-UAV test set. Standard and re-ranked Acc@0.5 use identical weights; $\Delta$ is the gain in percentage points. Each $\lambda^{*}$ is selected on held-out training data.}
\label{tab:rerank}
\renewcommand{\arraystretch}{1.08}
\centering
\small
\setlength{\tabcolsep}{4pt}
\begin{tabular}{@{}lcccc@{}}
\toprule
Configuration & $\lambda^{*}$ & Standard & Re-ranked & $\Delta$ \\
\midrule
Exemplar + alignment & 0.7 & 0.5159 & 0.5364 & +2.05 \\
\quad + reasoning & 0.7 & 0.5796 & 0.5888 & +0.92 \\
\quad + scene supervision & 0.5 & \textbf{0.6209} & \textbf{0.6255} & +0.46 \\
\bottomrule
\end{tabular}

\end{table}

\subsection{Ablation Studies}
\textbf{Cumulative contributions.}
Table~\ref{tab:chain} follows the method order, raising Acc@0.5 from 0.4104 to 0.6251. Each configuration uses one fixed seed and a shared evaluation runtime. These cumulative contrasts are not leave-one-out estimates.

\textbf{Reasoning and exemplar design.}
Original visual-first controls separate query-only derivation (0.5159 to 0.5556 Acc@0.5) from the role cue (0.5796). Satellite-conditioned regeneration and additive enrichment instead lose about 3.0 points relative to their visual base. Gated cross-attention outperforms additive and early satellite fusion (0.4651 versus 0.4580/0.4044). With alignment fixed, nine tokens outperform one or 49 (0.5159 versus 0.5091/0.4907; Table~\ref{tab:abl}).

\textbf{Scene supervision and re-ranking.}
Hard negatives lower their alignment base from 0.5091 to 0.4799 Acc@0.5; soft scene supervision raises its positionally cued base from 0.5796 to 0.6209. Different bases preclude a direct comparison. Re-ranking adds 2.05, 0.92, and 0.46 points across the visual, reasoning, and scene-supervised bases (Table~\ref{tab:rerank}). The selected $\lambda$ falls from 0.7 to 0.5 with scene supervision. Accuracy varies by at most 0.03 points for $K_r\in\{3,5,10\}$.

\textbf{Score aggregation.}
Mean aggregation (Eq.~\ref{eq:aggscore}) remains strongest: discriminability weighting changes Acc@0.5 by at most 0.06/0.01 points on reasoning-adapted/positionally cued bases; top-$k$ pooling loses 0.35--0.43 points. Learned token re-weighting reduces the visual base from 0.5159 to 0.5113. Further controls appear in the Appendix.




\section{Conclusion}

We presented GAV, a view-aware framework that rethinks cross-view object geo-localization through target-centered semantic reasoning and discriminative instance grounding. By integrating complementary linguistic and visual evidence, GAV mitigates viewpoint-dependent ambiguity and improves target discrimination among competing satellite instances. It achieves 62.55\% Acc@0.5 on MoP-UAV (+18.19 percentage points) and consistent improvements on CVOGL-Text.

\textbf{Limitations.}
GAV relies on offline description generation and annotated competitors, with only 11.2\% of CVOGL-Text training samples containing eligible competitors. The substantial contribution of query-position cues also raises questions about generalization to unseen settings. Generated expressions undergo sampled manual screening, and training comparisons use a single seed. Future work will explore online reasoning, annotation-efficient supervision, and improved cross-view robustness.

\clearpage
\section*{LLM Usage Statement}
We used OpenAI ChatGPT to assist with editing and polishing author-written text, refining the related-work citations, and editing author-written code. The authors take full responsibility for all submitted content.
{\small
\bibliographystyle{ieeenat_fullname}
\bibliography{references}
}
\end{document}